\documentclass{article}
\usepackage[accepted]{icml2018}

\usepackage{microtype}
\usepackage{graphicx}
\usepackage{xcolor}
\usepackage{booktabs} % for professional tables

\usepackage{times}
\usepackage{epsfig}
\usepackage{graphicx}
\usepackage{amsthm}
\usepackage{amsmath}
\usepackage{amssymb}
\usepackage{mathrsfs}
\usepackage[hidelinks]{hyperref}

\usepackage[utf8]{inputenc} % allow utf-8 input
\usepackage[T1]{fontenc}    % use 8-bit T1 fonts     % hyperlinks
\usepackage{url}            % simple URL typesetting
\usepackage{booktabs}       % professional-quality tables
\usepackage{amsfonts}       % blackboard math symbols
\usepackage{nicefrac}       % compact symbols for 1/2, etc.
\usepackage{microtype}      % microtypography

\usepackage[algo2e,ruled,vlined]{algorithm2e}

\usepackage{graphicx} 
\usepackage{subfigure} 
\usepackage{amsmath}
\usepackage{amsthm}
\usepackage{amssymb}
\usepackage{mathrsfs}
\usepackage{bm}
\usepackage{multirow}
\usepackage{multicol}

\theoremstyle{plain}

\usepackage{breqn}

\def\abovestrut#1{\rule[0in]{0in}{#1}\ignorespaces}
\def\belowstrut#1{\rule[-#1]{0in}{#1}\ignorespaces}

\SetNlSty{}{}{}

\let\oldnl\nl% Store \nl in \oldnl
\newcommand\nonl{%
  \renewcommand{\nl}{\let\nl\oldnl}}% Remove line number for one line

\def\1{{\bf{1}}}
\def\0{{\bf{0}}}

\newcommand{\beq}{\begin{equation}}
   \newcommand{\eeq}{\end{equation}}

\newcommand{\R}{\mathbb{R}}

\hypersetup{
colorlinks,
urlcolor=magenta,
linkcolor=black,
citecolor=black}
\icmltitlerunning{Deep $k$-Means: Re-Training and Parameter Sharing with Harder Cluster Assignments for Compressing Deep Convolutions}

\begin{document}

\twocolumn[
\icmltitle{Deep $k$-Means: Re-Training and Parameter Sharing with Harder Cluster Assignments for Compressing Deep Convolutions}



\icmlsetsymbol{equal}{*}

\begin{icmlauthorlist}
\icmlauthor{Junru Wu}{tamu}
\icmlauthor{Yue Wang}{rice}
\icmlauthor{Zhenyu Wu}{tamu}
\icmlauthor{Zhangyang Wang}{tamu}
\icmlauthor{Ashok Veeraraghavan}{rice}
\icmlauthor{Yingyan Lin}{rice}
\end{icmlauthorlist}

\icmlaffiliation{tamu}{Department of Computer Science and Engineering, Texas A\&M University, College Station, TX, USA;}
\icmlaffiliation{rice}{Department of Electrical and Computer Engineering, 
Rice University, Houston, TX, USA}

\icmlcorrespondingauthor{Zhangyang Wang}{atlaswang@tamu.edu}
\icmlcorrespondingauthor{Yingyan Lin}{yingyan.lin@rice.edu}

\icmlkeywords{Network Compression, Clustering, K-means, Energy-Efficient Learning}

\vskip 0.3in
]



\printAffiliationsAndNotice{}  % leave blank if no need to mention equal contribution


\begin{abstract}
The current trend of pushing CNNs deeper with convolutions has created a pressing demand to achieve higher compression gains on CNNs where convolutions dominate the computation and parameter amount (e.g., GoogLeNet, ResNet and Wide ResNet). Further, the high energy consumption of convolutions limits its deployment on mobile devices. To this end, we proposed a simple yet effective scheme for compressing convolutions though applying k-means clustering on the weights, compression is achieved through weight-sharing, by only recording $K$ cluster centers and weight assignment indexes.
We then introduced a novel spectrally relaxed $k$-means regularization, which tends to make hard assignments of convolutional layer weights to $K$ learned cluster centers during re-training. We additionally propose an improved set of metrics to estimate energy consumption of CNN hardware implementations, whose estimation results are verified to be consistent with previously proposed energy estimation tool extrapolated from actual hardware measurements. We finally evaluated Deep $k$-Means across several CNN models in terms of both compression ratio and energy consumption reduction, observing promising results without incurring accuracy loss. The code is available at \url{https://github.com/Sandbox3aster/Deep-K-Means}
\vspace{2em}
% \vspace{-1em}
\end{abstract}
% \url{https://github.com/Sandbox3aster/Deep-K-Means}
% \href{https://github.com/Sandbox3aster/Deep-K-Means}{\color{magenta}{https://github.com/Sandbox3aster/Deep-K-Means}}

% \hypersetup{
% colorlinks=false}

\section{Introduction}
Convolutional neural networks (CNNs) have gained considerable interest due to their record-breaking performance in many recognition tasks \cite{AlexNet,Object,DeepFace}. In parallel, there has been a tremendously growing need to bring CNNs into resource-constrained mobile devices in line with the recent surge of edge computing in which raw data are processed locally in edge devices using their embedded machine learning algorithms ~\cite{Edge} \cite{ISCAS_PredictiveNet}. The advantage lies in that local processing avoids transferring data back and forth between data centers and edge devices, thus reducing communication cost, latency, and enhancing privacy. 
%As such, many smartphones and handheld devices are integrated with intelligent user interfaces and applications, such as handwriting recognition, speech-based assistants (e.g. Siri), and face recognition enabled phone-unlock (e.g. FaceID). In addition to smartphones, CNNs are also expected to execute locally on a wider range of mobile and Internet-of-Thing (IoT) devices, such as wearables (e.g. Apple watch) and smart home infrastructure (e.g. Amazon Echo). 
However, deploying CNNs into resource-constrained platforms is a non-trivial task. Devices at the edge, such as smart phones and wearables, have limited energy, computation and storage resources since they are battery-powered and have a small form factor. In contrast, powerful CNNs require a large number of weights that corresponds to considerable storage and memory bandwidth. For example, the amount of weights in state-of-the-art CNNs AlexNet and VGG-16 are over 200MB and 500MB, respectively \cite{bib:arXiv2015:han}. 
%Therefore, it can be prohibitively difficult to deploy deep CNNs into resource-constrained platforms and 
Further, CNN-based applications can drain a battery very quickly if executed frequently. For example, smartphones nowadays cannot even run classification using AlexNet in real-time for more than one hour \cite{yang2017designing}. 

To close the gap between the constrained resources of edge devices and the growing complexity of CNNs, compression techniques have been widely investigated to reduce the precision of weights and the number of operations during or after CNN training in order to shrink their large implementation cost while maintaining the desired inference performance. Various CNN compression techniques have been proposed, such as weight compression~\cite{bib:bhattacharya2016:CD-ROM} \cite{bib:arXiv2015:han} \cite{bib:lane2016:IPSN} and decomposition~\cite{bib:ICLR2017:soravit} \cite{bib:arXiv2017:Howard} \cite{wang2015deepfont}, and compact architectures~\cite{bib:iandola2016:arxiv} \cite{bib:lin2013:NIN}. However, there are two major shortcomings in existing CNN compression techniques. 
\begin{itemize}
\vspace{-1em}
\item
A myriad of CNN compression techniques focus on the fully-connected layers of CNNs which conventionally have dominant parameters. However, recent successful CNNs tend to shift more parameters towards convolutional layers and have only one or even no fully-connected layers. For example, 85\% of the parameters lie in the convolutional layers of GoogleNet \cite{chen2016compressing}. Despite the growing trend toward CNN models using more convolutional layers and fewer fully-connected layers, only few compression techniques are dedicated for convolutional layers.
\vspace{-1.5em}
\item 
 The majority of CNN compression techniques, including most of the very few that focus on compressing convolutional layers, are designed to merely reduce the CNN model size or the amount of computation, which does not necessarily lead to reduced energy consumption. In fact, the recent work \cite{yang2017designing}
argues that the number of weights, multiply-and-accumulate (MAC) operations, and speedup ratio are often not good approximations for energy consumption, which also heavily depends on memory data movement and more. 
% argues that the number of weights, multiply-and-accumulate (MAC) operations are often not good approximations for energy consumption, it also heavily depends on memory data movement.
%  argues that neither the number of weights nor multiply-and-accumulate (MAC) operations reflect the actual energy consumption, memory accesses during data movement also plays a major role in the total energy cost. 
%  In fact, fetching data from the DRAM for an operation consumes orders of magnitude higher energy than the computation itself. 
For example, the authors show an interesting result that although SqueezeNet \cite{bib:iandola2016:arxiv} has 51.8$\times$ fewer weights than AlexNet, it consumes 33\% more energy due to its larger amount of computation and data movement. A compression technique that aims to reduce both model size and energy-aware complexity is hence highly desired for enabling extensive resource-constrained CNN applications. %to resource-constrained platforms. 
%\vspace{-1em}
\end{itemize}

\subsection{Contribution}
\vspace{-0.5em}
In this paper, we propose \textit{Deep $k$-Means}, a compression pipeline that is well suited for trimming down the complexity of convolutional layers that dominate both the model size as well as energy consumption of recently developed state-of-the-art CNNs. \textit{Deep $k$-Means} consists of two steps. First, a novel spectrally relaxed $k$-means regularization is developed to enforce highly clustered weight structures during re-training. After that, compression is performed via weight-sharing, by only recording cluster centers and weight assignment indexes. We evaluate the performance of \textit{Deep $k$-Means} in comparison with several state-of-the-art compression techniques focused on compressing convolutional layers. 
The results show that \textit{Deep $k$-Means} consistently achieves higher accuracy at the same compression ratio (CR) as its competitors. Furthermore, \textit{Deep $k$-Means} is also evaluated in terms of energy-aware metrics developed by us, and its compressed models show favorable energy efficiency as well. Our main contributions are summarized as follows:
\begin{itemize}
\vspace{-1em}
\item
We introduce a novel spectrally relaxed $k$-means regularization that automatically learns hard(er) assignments of convolutional layer weights during re-training, to favor the subsequent compression via $k$-means weight-sharing. Our regularization approach is effective, efficient, simple to implement and use, and easily scalable to large CNN models. 
\vspace{-0.5em}
\item
Inspired by a recently developed dataflow called ``row-stationary'', that minimizes data movement energy consumption on CNN hardware implementation \cite{Eyeriss-isca}, we reformulate the weights into row vectors for weigh-sharing clustering. 
%In addition to satisfactory compression rates achieved,
%compared to merely performing weight-sharing among vectorized weight filters, 
Such a formulation has the potential to result in CNN models that are in favor of energy-efficient hardware implementation. 
\vspace{-0.5em}
\item 
In order to bridge the gap between algorithm and hardware design of CNNs, we propose an improved set of energy-aware metrics based on \cite{pmlr-v70-sakr17a}. Our energy consumption estimation results are verified to be consistent with those from the tool in \cite{yang2017designing}, which was extrapolated from actual hardware measurements. We expect our metrics to broadly benefit future research in energy-aware CNN design. 
%\textcolor{red}{To YingyanL here need updating results, and emphasize more on energy results, e.g., consistency with MIT.} Notably, we apply \textit{Deep $k$-Means} to a state-of-the-art deep CNN, Wide ResNet \cite{zagoruyko2016wide}, observing only 0.51\% loss of top-1 accuracy at 45 times compression.
\vspace{-0.5em}
\end{itemize}
%Experimental results show that the \textit{Deep $k$-Means} outperforms existing competitive methods in terms of both compression rate and our energy-aware performance metrics. 

%In the rest of this paper, we review related work in Section 2, describe the proposed technique in Section 3, and thoroughly evaluate the performance of our technique over different CNN models and in comparison of state-of-the-art convolutional layers compression techniques in Section 4. Finally, we conclude this work in Section 5.
\vspace{-0.5em}

\subsection{Related Work}
\vspace{-0.5em}
%Substantial research efforts have been made to compress deep neural networks, and our work is closely related to the following research.
%\subsection{Parameter Pruning and Sharing}
Parameter pruning and sharing has been used both to reduce network complexity and to avoid over-fitting. Early pruning approaches include Biased Weight Decay \cite{NIPS1988_156}, Optimal Brain Damage \cite{Cun:1990:OBD:109230.109298}, and Optimal Brain Surgeon \cite{NIPS1992_647}. Recent works \cite{DBLP:journals/corr/SrinivasB15} made use of the redundancy among neurons. The Deep Compression method introduced
in \cite{bib:arXiv2015:han} employed a three stage pipeline to prune the redundant connections, quantize the weights via scalar weight sharing, and then encode the quantized weights using Huffman coding. An effective soft weight-sharing method described  in \cite{ullrich2017soft} showed competitive CRs on state-of-the-art CNNs, e.g., Wide ResNet. 

With fully-connected layers traditionally considered as the memory bottleneck, 
numerous works focused on compressing these layers. For example, \cite{DBLP:journals/corr/GongLYB14} proposed applying k-means clustering to the densely-connected layers and showed a good balance between model size and accuracy. \cite{DBLP:journals/corr/ChenWTWC15} proposed HashedNet that used a low-cost hash function to group weights into hash buckets for parameter sharing. On the other hand, a few recent works embraced the trend towards more convolutional layers in CNNs and attempted to compress convolutional layers. For example, \cite{chen2016compressing} proposed an architecture called FreshNets to compress filters of convolutional layers in the frequency domain.
%using discrete cosine transform and then randomly group the resulting frequency parameters into buckets
The recent work \cite{abbasi2017structural} iteratively pruned filters based on the
classification accuracy reduction index, and achieved substantially higher classification accuracy compared to other structural compression schemes, e.g., \cite{He,DBLP:journals/corr/LiKDSG16}.

\section{Proposed Approach}

\subsection{Parameter Sharing via Row-wise $k$-Means}
\label{testing}
\vspace{-0.5em}
Assuming a convolutional layer $\in \R^{s \times s \times c \times m}$, where $s$ denotes the filter size, $c$ the input channel number, and $m$ the output channel number. Following the convention in CNNs, we reshape it as a matrix $W \in \R^{s \times N}$, where $N = s \times c \times m$, each column vector $\in \R^{s}$ being \underline{a row} from an original convolutional filter. Following the product quantization approach for fully-connected layers in \cite{gong2014compressing}, we treat all columns of $W$ as $N$ samples, and apply $k$-means to assign them with $K$ clusters. When $K \ll N$, we need only to store the cluster indexes and codebooks after $k$-means. We define the \textit{cluster rate} for each layer as $\frac{K}{N}$ here.
%leading to a compression rate of $32sN/(32KN + s\log_2 K)$, if we assume 8 bits per index. 
For compressing multiple convolutional layers, we adopt a ``uniform parameter sharing'' scheme for simplicity, i.e., each convolutional layer chooses its $K$ value such that all layers have the same cluster rate, expect for the first layer whose cluster rate is often set higher.

%will perform the above independently for each layer with the same $K$.

We notice other alternatives to enforce structured parameter sharing among convolutional layers using $k$-means, e.g., reshaping each convolutional filter as vectors $R^{s^2}$ and then clustering over $c \times m$ samples, or converting each output channel into $\R^{s \times s \times c}$ and clustering over resulting $m$ samples. In practice, we find their performance to be close (with $k$ chosen in different proper ways).
%and the above row-wise $k$-Means to converge fastest among all. 
One major motivation for choosing the row-wise $k$-means is that it could lead to higher data reuse opportunity and thus result in more energy-efficient hardware implementations, according to the row-stationary dataflow recently proposed in \cite{Eyeriss-isca}, which has shown to be superior in terms of energy efficiency compared to other dataflows. Another motivation arises from reducing the complexity (see Section \ref{train}).
%complexity part. 

\vspace{-0.5em}
\subsection{$k$-Means Regularized Re-Training}
\label{train}
\vspace{-0.5em}
Simply pruning or sharing weights in CNNs will usually hurt the inference accuracy. Re-training has often been exploited to enforce the favorable structures in the pruned/shared weights and compensate for the accuracy loss \cite{bib:arXiv2015:han}. In order to be compatible with $k$-means parameter sharing, we would favor a re-training scheme that ``naturally'' encourages the weights to be concentrated tightly around, or exactly at, a number of cluster components which are optimized for high predictive accuracy. The goal is fulfilled by introducing a novel \textit{spectrally relaxed $k$-Means regularization} below. 

The original sum-of-squares function of $k$-Means usually employs a Lloyd-type algorithm to solve. The spectral relaxation technique of $k$-Means was introduced in \cite{zha2002spectral}, by first equivalently re-formulating sum-of-squares into a trace form with special constraints. Specifically, to cluster $N$ samples of $\R^s$, represented as $W \in \R^{s \times N}$, into $K$ clusters, the spectral relaxation converts the $k$-means objective into the following problem:
\begin{equation}
 \begin{aligned}
\min_{W; F \in \mathcal{F}}\, Tr(W^T W) - Tr(F^TW^TWF),
  \end{aligned}
 \label{original}
\end{equation}
where $Tr$ denotes the matrix trace. $F \in \R^{N \times k}$ is the normalized cluster index matrix, and $\mathcal{F}$ denotes its special structure requirement: $F_{ij} = 1/\sqrt{n_j}$ if column $i$ belongs to the cluster $j$ and there is a total of $n_j$ samples in the cluster $j$; and $F_{ij} = 0$ otherwise, $i = 1, ..., N$, $j = 1, ..., K$, and $\sum_{j = 1}^{K} n_j = N$. The original spectral relaxation \cite{zha2002spectral} considers $W$ as given; thus (\ref{original}) is reduced to:   
\begin{equation}
 \begin{aligned}
\max_{F \in \mathcal{F}}\, Tr(F^TW^TWF)
  \end{aligned}
 \label{original2}
\end{equation}
The authors of \cite{zha2002spectral} then proposed ignoring the special structure of $F$ and let it be an arbitrary orthogonal matrix. (\ref{original2}) is thus relaxed to a trace maximization problem over a Stiefel manifold:
\begin{equation}
 \begin{aligned}
\max_{F}\, Tr(F^TW^TWF), \, s.t.\,\, F^TF = I
  \end{aligned}
 \label{original3}
\end{equation}
It results in a closed-form solution of $F$, by composing the first $k$ singular vectors of $W$, according to the well-known Ky Fan theorem. 

As a \underline{critical difference} with \cite{zha2002spectral}, here our goal is not to cluster a \textit{static} $W$. Rather, we would like to encourage $W$ to stay ``suited'' for $k$-means during the dynamic re-training, without incurring a significant increase in complexity. We are thus motivated to utilize (\ref{original}) as a regularization term on learning $W$, rather than a stand-alone objective. We discuss just one convolutional layer $W$ for simplicity: assume that the original CNN training minimizes the energy function $E(W)$, w.r.t. $W$. The retraining minimizes the regularized objective below ($\lambda$ is a scalar):
\begin{equation}
 \begin{aligned}
\min_{W, F}\,\, & E(W) + \frac{\lambda}{2} [Tr(W^T W) - Tr(F^TW^TWF)], \\ 
s.t.\, & F^TF = I
  \end{aligned}
 \label{retrain}
\end{equation}
Note that $F$ is treated as an auxiliary variable to promote a clustered structure in $W$. Solving (\ref{retrain}) could be iterated between the updates of $W$ and $F$. Updating $W$ can follow the standard stochastic gradient descent (SGD), with the gradient given as: $\nabla E(W) + \lambda W (I - FF^T)$. $F$ is updated using the same closed-form solution to (\ref{original3}), by computing the $k$-truncated singular value decomposition (SVD) of $W$. 

%Remember that the spectrally relaxed $k$-means regularization term $\frac{\lambda}{2} [Tr(W^T W) - Tr(F^TW^TWF)]$ is directly derived from the original ``hard'' $k$-means objective (\ref{original}). 

By the interaction between $F$ and $W$ during re-training, the regularization keeps $W$ in a highly clustered state, in addition to optimizing it for inference accuracy. Although $F$ has been relaxed from the ``hard'' normalized cluster index matrix to an arbitrary orthogonal one, we observe in practice that it still tends to enforce weights close to those taking around $K$ unique vector values, i.e., encouraging ``approximately hard'' (or ``harder'' than soft weight sharing) $K$-cluster assignments during re-training.

%relaxed 
%Since it is essentially derived from the original ``hard'' $k$-means objective (\ref{original}), it will tend to enforce weights to take strictly no more than $K$ unique vector values during re-training.

In \textit{Deep $k$-Means}, starting from an uncompressed pre-trained model as initialization, we will re-train it with adding this novel data-dependent weight regularizer (\ref{retrain}) to each convolutional layer, while other training protocols remain unchanged. The re-training typically converges into a much smaller number of epochs than in the original training. After that, we apply row-wise $k$-means on the learned $W$ for the final parameter-sharing step.

\vspace{-1em}
\paragraph{Complexity Analysis}
For each convolutional layer $W$, the extra complexity incurred by applying the spectrally relaxed $k$-means regularization term includes two parts: (1) updating $W$: the only extra burden is to compute $\lambda W (I - FF^T)$, which takes $\mathcal{O}(sKN)$ or $\mathcal{O}(s^2cmK)$ (computing $WFF^T$); (2) updating $F$ via SVD, which costs $\mathcal{O}(s^2N)$ or $\mathcal{O}(s^3cm)$: that also serves another motivation to create $W$ with lower row dimensions (e.g., $s$ rather than $s^2$ or $s^2c$), since it will reduce the SVD complexity of $W$. Considering that $s$ is usually small, the total extra complexity $\mathcal{O}((s^2K + s^3)cm)$ is quite affordable, enabling our methods to scale well for modern CNNs. In practice, we also implement the $F$ update in a very ``lazy'' way so that SVD will merely be computed once for every five epochs, for further accelerating the re-training, with only marginal impacts on the result.

%We next solve a $(k-1)$-cluster problem form of (\ref{retrain}) on the remaining sub-matrix $\in \R^{s \times (1-p)N}$.  

% Another difference with \cite{ullrich2017soft} and other methods is that in this paper the approach does not try to enforce values to a cluster center in 0. This can increase the compression rate but reduce the possible gain in performance/speed due to removing 0 weights. The authors should at least mention this and preferably test with or without enforcing a cluster centered in 0.

%\footnote{For example, without sparsity promotion, our Wide ResNet compression achieved 0.51\% top-1 loss, while it becomes 1.63\% after adding sparsity. However, it still outperforms the state-of-the-art.}

%ICLR 17: ``However, for large networks such as VGG with 138 M parameters the algorithm is too slow to get usable results.''
\vspace{-0.5em}
\subsection{Comparison with Existing Work} 
\vspace{-0.5em}
Directly enforcing a $k$-means friendly weight structure is not straightforward. \cite{ullrich2017soft} presents an elegant and inspiring Bayesian regularization form of ``soft cluster assignment''. During re-training, the authors fit a Gaussian mixture model (GMM) prior model over the weights, to encourage the distribution of weights to be close to $K$ clusters. After re-training, each weight was quantized to the mean of the GMM component that takes most responsibility, for parameter sharing. Their pipeline is the closet peer work to ours, with the major difference being that we pursue harder cluster assignment during re-training. As is well known, GMM is reduced to $k$-means when the mixture variance gets close to zero. Therefore, the retraining process in \cite{ullrich2017soft} could also be viewed as a ``softened'' version of $k$-means. However, the differences between the two methods manifest in multiple folds:
\begin{itemize}
\vspace{-1em}
\item First, 
%both \textit{Deep $k$-Means} and \cite{ullrich2017soft} achieve parameter-sharing in the same ``hard'' clustering way. 
our ``harder'' cluster assignment is directly derived from the original $k$-means objective (\ref{original}). We expect it to be \underline{better aligned} with the $k$-means parameter sharing stage. Our experimental observations show that this leads to more skewed weight distributions, and achieves better results than \cite{ullrich2017soft}. 
%during the re-training stage to be \underline{better aligned} with the parameter sharing stage. 
\vspace{-0.5em}
\item Second, compared to the Bayesian form in \cite{ullrich2017soft}, our regularization adds very little extra complexity to the standard SGD. The implementation only calls for minor changes (a new regularizer term); and thanks to its low complexity, it is ready to be applied to larger-scale CNNs. 
\vspace{-0.5em}
\item Third, \cite{ullrich2017soft} discussed their high sensitivity to the choices of learning rates for mixture parameters (e.g., means, log-variances): a higher learning rate may cause model collapse and a lower one results in slow convergence. In contrast, \textit{Deep $k$-Means} has merely one hyper-parameter $\lambda$. 
We find \textit{Deep $k$-Means} insensitive to $\lambda$ ($\lambda$ between $10^{-4}$ and $10^{-3}$ is found to work almost equally well). 
%It also works well with the same default $p$ for all models: 0 for the first layer, and 0.5 for all remaining layers. The only empirical exception is Wide ResNet, as we will specify next. Beyond that, 
\textit{Deep $k$-Means} needs no special learning rate scheduling. It is also free of postprocessing, e.g., removing redundant components as \cite{ullrich2017soft} needed to.
\vspace{-0.5em}
% \item Finally, we observe the notable performance advantage of \textit{Deep $k$-Means} over \cite{ullrich2017soft}, when both are applied to compressing the state-the-art CNNs: we compare their results on Wide ResNet in Section 4.
%\vspace{-0.5em}
\end{itemize}

%no obvious computational bottleneck and are ready to generalize larger networks such as VGG \cite{}, as part of our ongoing work.

%similar to ICLR'17: MDL principle

\section{Energy-Aware Metrics for CNN Energy Consumption Estimation}
%\textcolor{red}{To Yingyan: I created a new paragraph to emphasize novelty, and to avoid too long experiment part. Please revise}

While CR or reduction in the number of operations are widely adopted by existing CNN compression techniques as generic performance metrics, these metrics are not necessarily tied to improved energy efficiency as pinpointed by \cite{yang2017designing} according to their energy estimation tool extrapolated from actual hardware measurements. Therefore, it is important to evaluate compression techniques using a set of energy-aware metrics other than CR. 

However, it is non-trivial to estimate the energy consumption of CNNs because a significant portion of energy consumption in CNNs is consumed by data movement, which mainly depends on the employed memory hierarchy and dataflow when implementing CNNs and is thus difficult to be estimated directly from the model. An energy estimation tool extrapolated from actual hardware measurements was proposed by \cite{yang2017designing} to bridge the gap between algorithm and hardware design. Unfortunately, their tool currently only 
supports AlexNet and GoogLeNet_v1. 

We hereby propose the following energy-aware metrics: %which were first proposed in \cite{pmlr-v70-sakr17a}:
\begin{itemize}
\vspace{-1em}
\item 
\textbf{Computational cost} measures the computational resources needed to generate a single decision and is defined in terms of the number of 1 bit full adders (FAs), which is a canonical building block of arithmetic units. Specifically, assuming that the arithmetic operations are executed using the commonly employed ripple carry adder and BaughWooley multiplier architectures, the number of FAs needed to compute a $D$-dimensional dot product between the activations and weights is \cite{RD-SEC}:
\begin{eqnarray}
DB_{\mathbf{w}}B_{\mathbf{x}}+(D-1)(B_{\mathbf{x}}+B_{\mathbf{w}}+\left\lceil \log_{2}(D)\right\rceil -1)\label{eq: computational cost}
\end{eqnarray}
where $B_{\mathbf{w}}$ and $B_{\mathbf{x}}$ denote the fixed-point precision assigned to the weights and activations, respectively. 
\vspace{-0.5em}
\item 
\textbf{Weight representational cost} measures the storage
complexity and data movement costs corresponding to the weights and is defined as the product between the total number of bits needed to represent all weight parameters and the total number of times that the weights are used to compute convolutions: 
\begin{eqnarray}
N_{\mathbf{w}}\left|\mathcal{W}\right|B_{\mathbf{w}}\label{eq: weight representational cost}
\end{eqnarray}
where $N_{\mathbf{w}}$ and $\mathcal{W}$ denote the total number of times that the weights are used to compute convolutions and the index sets of all weights in the network, respectively. 
\vspace{-0.5em}
\item 
\textbf{Activation representational cost} is similar to the weight representational cost above and is defined as: 
\begin{eqnarray}
N_{\mathbf{x}}\left|\mathcal{X}\right|B_{\mathbf{x}}\label{eq: activation representational cost}
\end{eqnarray}
$N_{\mathbf{x}}$ and $\mathcal{X}$ denote the total number of times that the activations are used to compute convolutions and the index sets of all activations in the network, respectively.\vspace{-0.5em}
\end{itemize}
The concepts of computational and representational costs were first proposed in \cite{pmlr-v70-sakr17a} to describe network complexity. To better reflect the CNN energy cost, we modify the definition of representational cost in \cite{pmlr-v70-sakr17a} to include the number of times that weights or activations are loaded for computing convolutions, in order to capture the associated data movement cost. Specifically, if a certain weight filter is removed due to compression, then the corresponding activation would be loaded less frequently, thus leading to reduced data movement costs. This can be reflected by the reduction of $N_{\mathbf{x}}$ in our modified definition but not the originally defined representational cost in \cite{pmlr-v70-sakr17a}. 
%as the latter is to describe the network complexity. 
Also, we separate the representational costs for the weights and activations to evaluate the impact of compression in more detail. 
\vspace{-0.5em}
\section{Experiments}
\vspace{-0.3em}
We evaluate \textit{Deep $k$-Means} %for compressing CNNs whose convolution layers dominate the total parameter amounts, 
in terms of CR and energy-aware metrics respectively, with the resulting accuracy loss $\Delta$ after compression, using two sets of experiments. The default $\lambda$ is $10^{-4}$. Unless otherwise specified, we will focus on compressing convolutional layers only. 

\underline{For the first set of experiments} on CR, we first create a simple baseline CNN for simulation experiments w.r.t. varying CR. The CR definition follows \cite{bib:arXiv2015:han}. We then compare \textit{Deep $k$-Means} against four latest and competitive comparison baselines for compressing convolutional layers: Ultimate Tensorization \cite{garipov2016ultimate}, FreshNet \cite{chen2016compressing}, Greedy Filter Pruning \cite{abbasi2017structural}, and Soft Weight-Sharing \cite{ullrich2017soft}. The first three have been optimized towards compressing CNNs dominated by the convolutions and reported results on their self-designed models. \cite{ullrich2017soft} outperformed strong baselines such as \cite{DBLP:journals/corr/HanPTD15} on the standard MNIST benchmark; the authors then reported compression results on the state-of-the-art Wide ResNet model \cite{zagoruyko2016wide} that mainly consist of convolutions. We also compare \textit{Deep $k$-Means} with the baseline of \textit{Deep $k$-Means} without re-training (\textit{Deep $k$-Means WR}), i.e., directly performing row-wise $k$-means on original weights. 
%The compression ratio (CR) is defined the same as \cite{bib:arXiv2015:han}. 

\underline{For the second set of experiments}, we first validate the estimated energy consumption using our metrics to match the actual hardware-based extrapolation \cite{yang2017designing}, and then evaluate \textit{Deep $k$-Means} against the aforementioned baselines from an energy consumption perspective. While it is overall challenging to estimate energy consumption accurately due to the
multitude of factors involved, our proposed metrics are simple, effective (i.e., showing a good match with the results extrapolated by actual hardware measurement), and thus can help CNN model designers understand various design trade-offs. We also provide insights regarding the impact of different types (i.e., parameter-sharing or pruning) of compression techniques on the computational and representational costs in Eqs. \eqref{eq: computational cost}, \eqref{eq: weight representational cost} and \eqref{eq: activation representational cost}.

%- how to computer energy?
%- what you choose for compare, and why...

%Deep $k$-Means WR could be viewed extending Facebook method to conv

% \begin{figure}[htbp]
% \centering
% \begin{minipage}{0.45\textwidth}
% \centering{
% \includegraphics[width=\textwidth]{./images/ablation.eps}
% }\end{minipage}
% \caption{Accuracy after compression using \textit{Deep $k$-Means} and \textit{Deep $k$-Means WR}, w.r.t. different CRs.}
% %\vspace{-1em}
% \label{ablation}
% \end{figure}

\vspace{-0.5em}
\subsection{Comparison on Compression Ratio}

% \subsubsection{Simulation with varying CRs}

% We first perform a simulation experiment using a modified LeNet model from LeNet_Caffe\footnote{\url{https://github.com/BVLC/caffe/blob/master/examples/mnist/lenet.prototxt}}, in which we merge the two fully-connected layer to one $800 \times 10$ layer to make convolutional layer weights dominant. The modified LeNet provides a baseline (uncompressed) accuracy of 98.97\% on the MNIST dataset. We choose a range of CRs between 93\% and 99\% following the uniform parameter sharing scheme (see Section \ref{testing}). Figure \ref{ablation} depicts the accuracy-CR curves of both Deep $k$-Means and Deep $k$-Means WR. It is clearly observed that parameter-sharing with $k$-means alone (Deep $k$-Means WR) is fragile under high CRs and degrades the accuracy quickly. In comparison, Deep $k$-Means gains consistent advantages at the same CR, and the re-training stage proves to effectively restores accuracy loss: there is no visible accuracy loss until CR $\approx$ 98\%.

%\vspace{-0.6em}
\subsubsection{Comparison with Ultimate Tensorization}

\begin{table}[htp!]
\centering
\vspace{-1em}
\begin{tabular}{lll}
    \toprule
    Model    & $\Delta$ (\%) & CR  \\
    \midrule
    TT-conv (naive) & -$2.4$ & $2.02$ \\
    TT-conv (naive) & -$3.1$ & $2.90$ \\
    \hline
    TT-conv & -$0.8$  & $2.02$   \\
    TT-conv & -$1.5$ & $2.53$  \\
    TT-conv & -$1.4$ & $3.23$   \\
    TT-conv & -$2.0$ & $4.02$  \\
    \hline
%    Deep $k$-Means WR& +$0.05$ & 2 \\
%    Deep $k$-Means WR& -$0.08$ & 4 \\
    Deep $k$-Means & +$0.05$ & 2\\
    Deep $k$-Means & -$0.04$ & 4\\
    \bottomrule
  \end{tabular}
\caption{Compressing TT-conv-CNN in \cite{garipov2016ultimate}.}. 
\vspace{-2.5em}
\label{ttconv}
\end{table}

\cite{garipov2016ultimate} proposed a tensor factorization based method specifically for compressing convolutional layers. The authors proposed a Tensor Train (TT) Decomposition approach for convolutional kernel, denoted as \textit{TT-conv (naive)}. It could be further enhanced by introducing a new type of TT-conv layer, denoted as \textit{TT-conv}. The authors evaluated TT-conv (naive) and TT-conv on a self-designed architecture, called TT-conv-CNN, consisting of six convolutional layers and one fully-connected layer. TT-conv-CNN is dominated by the convolutions (occupying 99.54\% parameters of the network), and the authors reported the uncompressed model's top-1 accuracy of 90.7\% on CIFAR-10 \cite{krizhevsky2009learning}. 

We evaluate \textit{Deep $k$-Means} on the TT-conv-CNN model at CR = 2 and 4.  Table \ref{ttconv} compares them with the compression results in \cite{garipov2016ultimate}. \textit{Deep $k$-means} incurs minimal accuracy loss even at CR = 4. More surprisingly, it even slightly increases the accuracy after compression at CR = 2. It concurs with the previous observations by \cite{ullrich2017soft,cheng2017survey}: removing parameter redundancy improves CNN generalization on some small networks.

%e realize that sometimes compression might in turn slightly increases the accuracy rather than causing loss, such as reported by \cite{ullrich2017soft} on some small networks. Researchers have found that pruning unnecessary filters improves CNN generalization \cite{cheng2017survey}. Hence we also plan to explore whether \textit{Deep $k$-Means} could regularize CNNs to generalize better and gain more robustness.

%\textit{Deep $k$-Means} show clear superiority at CR = 2 and 4.

%Even \textit{Deep $k$-Means WR} shows clear margins over the TT-conv (naive) and TT-conv methods. Further boosts can be obtained for \textit{Deep $k$-Means} when re-training is done, leading to 0.17\% accuracy improvement at CR = 2 and 0.14\% at CR = 4.

\subsubsection{Comparison with FreshNet}
The Frequency-Sensitive Hashed Nets (FreshNets) was proposed in \cite{chen2016compressing} to exploit inherent redundancy in convolutional layers. The authors observed that convolutional weights to be typically smooth and low-frequency. They were thus motivated to first convert filter weights to the frequency domain, after which they group frequency parameters into hash buckets to achieve parameter sharing. The authors evaluated their method on their self-designed CNN (referred to as \textit{FreshNet-CNN} hereinafter) consisting of five convolutional layers and one fully-connected layer. They reported the uncompressed FreshNet-CNN to obtain the top-1 accuracy of 85.09\% on CIFAR-10. 

%(used for their CR = 16 experiment)/85.63\% (used for their CR = 64 experiment)

Table \ref{FreshNet} reports the compression results of \textit{Deep $k$-Means} and \textit{Deep $k$-Means WR} on FreshNet-CNN at CR = 16. We also include two original baselines in \cite{chen2016compressing}: low-rank decomposition (LRD) \cite{denil2013predicting} and HashedNet \cite{DBLP:journals/corr/ChenWTWC15}. In this example, even the accuracy of \textit{Deep $k$-Means WR} is very competitive. After re-training, \textit{Deep $k$-Means} shows a sharp further improvement.

% performs much worse than FreshNet under both CRs; its accuracy actually dramatically deteriorates at CR = 64. However, it is encouraging to observe that after re-training, 

% \textit{Deep $k$-Means} shows strong robustness to the severe information loss under high CRs, and stands the best among all methods under both CRs. 

% see the great help of re-training at CR = 64

\begin{table}[htp!]
\centering
\begin{tabular}{lll}
    \toprule
    Model    & $\Delta$ (\%) & CR  \\
    \midrule
    LRD & -$8.32$ & $16$ \\
    HashedNet & -$9.79$ & $16$ \\
    FreshNet & -$6.51$ & $16$ \\
    Deep $k$-Means WR& -$5.95$  & $16$ \\
    Deep $k$-Means & -$1.30$ & $16$ \\
%     \hline
%     LRD & -$19.98$ & $64$ \\
%     HashedNet & -$28.71$ & $64$ \\
%     FreshNet & -$16.42$ & $64$ \\
%     Deep $k$-Means WR&  -$66.92$ & $64$ \\
%     Deep $k$-Means &  -$7.47$ & $64$ \\
    \bottomrule
  \end{tabular}
\caption{Compressing FreshNet-CNN in \cite{chen2016compressing}.}. 
\vspace{-1em}
\label{FreshNet}
\end{table}

\vspace{-1em}
\subsubsection{Comparison with Greedy Filter Pruning}
\label{greedy_section}
The recent work \cite{abbasi2017structural} introduced a greedy structural compression scheme that prunes redundant convolutional filters in a trained CNN, based on a classification accuracy reduction (CAR) algorithm. The authors reported promising results on LeNet-5, AlexNet and ResNet-50, and their evaluations adopted a unique layer-wise compression fashion: taking LeNet-5 for example, each time the authors pruned filters in one convolutional layer (first or second) while leaving other layers untouched, and then reported the overall accuracy. 

\begin{figure}[htbp]
\vspace{-0.8em}
\centering
\begin{minipage}{0.4\textwidth}
\centering\subfigure[Comparison in the first convolutional layer]{
\includegraphics[width=\textwidth, height=0.18\textheight]{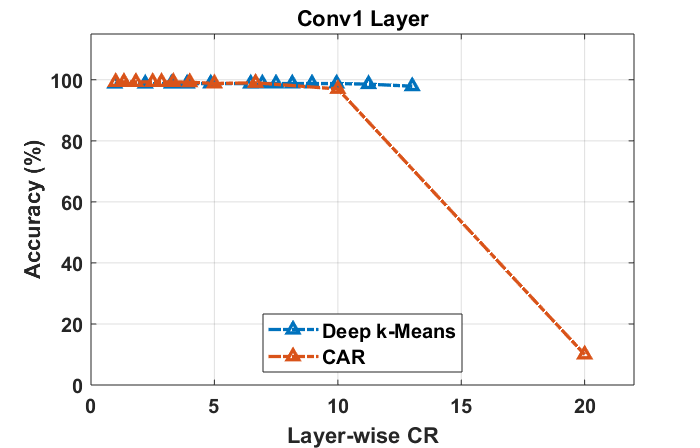}
}\end{minipage}
\begin{minipage}{0.4\textwidth}
\centering\subfigure[Comparison in the second convolutional layer]{
\includegraphics[width=\textwidth, height=0.18\textheight]{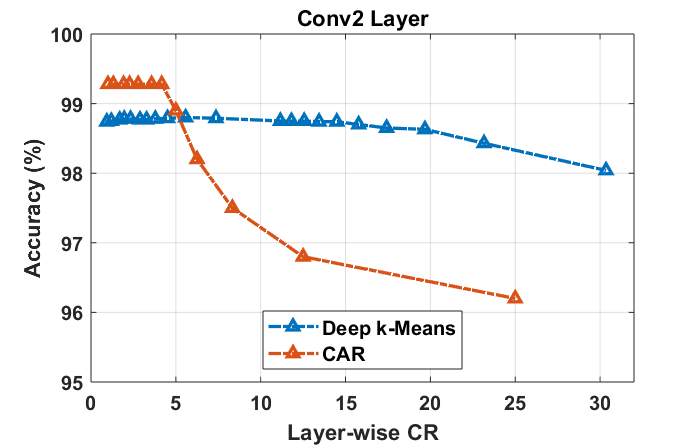}
}\end{minipage}
\vspace{-0.5em}
\caption{Compressing LeNet following the layer-wise setting in \cite{abbasi2017structural}: (a) The overall classification accuracy of LeNet when only the first convolutional layer (Conv1) is compressed, w.r.t. layer-wise CR; (b) The overall classification accuracy of LeNet when only the second convolutional layer (Conv2) is compressed, w.r.t. layer-wise CR.}
\vspace{-1em}
\label{greedy}
\end{figure}

We compare \textit{Deep $k$-Means} with CAR (with re-training, the best performer in \cite{abbasi2017structural}) using LeNet-5 on MNIST, and follow their layer-wise compression setting. Thus, unlike our other experiments, we report the accuracy w.r.t. ``layer-wise'' CR, i.e., measuring how many times the current layer is compressed, rather than the overall CR that measures the entire model. As Figure \ref{greedy} shows, both \textit{Deep $k$-Means} and CAR produce similar results at small layer-wise CRs; CAR is more competitive at small CRs for Conv2. However, \textit{Deep $k$-Means} is clearly superior at high layer-wise CRs for both layers.

% where their accuracies are almost identical to the uncompressed model. \textit{Deep $k$-Means} is more clearly superior to both CAR and \textit{Deep $k$-Means WR} at high layer-wise CRs, especially when compressing the first layer. 

% Stacked Configure
% Side by Side Configure
% \begin{figure*}[htbp]
% \centering
% \begin{minipage}{0.42\textwidth}
% \centering\subfigure[]{
% \includegraphics[width=\textwidth]{./images/conv1_retrain_compression_rate_new.png}
% }\end{minipage}
% \begin{minipage}{0.42\textwidth}
% \centering\subfigure[]{
% \includegraphics[width=\textwidth]{./images/conv2_retrain_compression_rate_new.png}
% }\end{minipage}
% \vspace{-1em}
% \caption{Compressing LeNet following the layer-wise setting in \cite{abbasi2017structural}: (a) The overall classification accuracy of LeNet when only the first convolutional layer (Conv1) is compressed, w.r.t. layer-wise CR; (b) The overall classification accuracy of LeNet when only the second convolutional layer (Conv2) is compressed, w.r.t. layer-wise CR.}
% \vspace{-0.5em}
% \label{greedy}
% \end{figure*}

%\footnote{LeNet-5 is a four-layer CNN consisting of two convolutional layers and two fully-connected layers, and its fully-connected layers cost considerable amounts of parameters. Therefore, the CRs in Figure \ref{greedy} are not as high as in other experiments.}.

%Performance of compression for LeNet

%LeNet 5: not dominated by convolutions, but \cite{abbasi2017structural} focuses on pruning filters in convolutional layers, belong to a different scheme. We keep Deep $k$-Means in convolutional layers too, thus making comparable baselines

\subsubsection{Comparison with Soft Weight Sharing}

\begin{table}[htp!]
\centering
\begin{tabular}{lll}
    \toprule
    Model    & $\Delta$ (\%) & CR  \\
    \midrule
    Soft Weight-Sharing & -$2.02$ & $45$ \\
    \hline
    %Deep $k$-Means WR & -$53.35$ & $40$ \\
    Deep $k$-Means WR & -$16.02$ & $45$ \\
    Deep $k$-Means WR & -$25.45$ & $47$ \\
    Deep $k$-Means WR &  -$45.08$ & $50$ \\
    \hline
    %Deep $k$-Means & -$1.42$ & $40$ \\
    %Deep $k$-Means (0.5)  &   -$3.76$ & 45 \\
    Deep $k$-Means & -$1.63$ & $45$ \\
    Deep $k$-Means & -$2.23$ & $47$ \\
    Deep $k$-Means &  -$4.49$ & $50$ \\
    \bottomrule
  \end{tabular}
\caption{Compressing Wide ResNet in comparison to soft weight-sharing \cite{ullrich2017soft}.}. 
\vspace{-2em}
\label{WRN}
\end{table}

% \begin{figure}[htbp]
% \centering
% \begin{minipage}{0.38\textwidth}
% \centering\subfigure[]{
% \includegraphics[width=\textwidth]{images/conv1_retrain_computational_cost.png}
% }\end{minipage}
% \begin{minipage}{0.38\textwidth}
% \centering\subfigure[]{
% \includegraphics[width=\textwidth]{images/conv2_retrain_computational_cost.png}
% }\end{minipage}
% \vspace{-0.5em}
% \caption{Energy Consumption Comparison using \textit{Deep $k$-Means}.}
% \vspace{-2em}
% \label{greedy}
% \end{figure}

% 93.82% uncompressed in our case

\cite{ullrich2017soft} reported the compression performance of soft weight-sharing on the state-of-the-art Wide ResNet model \cite{zagoruyko2016wide}, a convolution-dominant CNN with 2.7M parameters, at one single CR = 45 using CIFAR-10 (the uncompressed baseline top-1 error is 6.48\%). Thanks to the light computational burden of \textit{Deep $k$-Means}, we are able to evaluate various CRs. Note that at the same CR, soft weight-sharing and \textit{Deep $k$-Means} will lead to identical layer-wise dimensions and the same number of unique weights in each layer. Thus, their performance difference can only arise from the effects of their different regularization ways during re-training.

%\vspace{-1em}
\textit{\underline{Promoting Sparsity in Re-Training}.} During the review stage, \textit{one anonymous reviewer} commented that \cite{ullrich2017soft} tried to explicitly enforce weight values to a cluster centered at zero, while the above default routine of \textit{Deep $k$-Means} had not such constraint. 
Such a sparsity-promotion operation may marginally decrease compression performance as it restricts the flexibility of setting centroids, but can gain more in both speedup and energy savings \cite{parashar2017scnn}. To ensure a fair comparison with \cite{ullrich2017soft}, we implement a similar sparsity-promoting feature for \textit{Deep $k$-Means}, \textit{in this specific experiment only}. Without referring to sophisticated options such as semi-supervised clustering \cite{basu2002semi}, we follow a simple heuristic which incurs almost no extra complexity: at each time of ``lazy update'' for layer $W \in \R^{s \times N}$, we first rank all $N$ columns of $W$ in terms of their $\ell_2$ norms. We then assign the $pN$ (0 < $p$ <1) smallest-norm columns to one cluster with a fixed center at zero, before solving (\ref{retrain}). At the parameter-sharing step, we similarly threshold $pN$ smallest-norm columns in $W$ to be all-zero, and then perform $(k-1)$-clustering for remaining columns. The group of layer-wise $p$ that we used for all 16 layers is: [0, 0.3, 0.4, 0.5, 0.4, 0.4, 0.5, 0.5, 0.5, 0.5, 0.5, 0.6, 0.9, 0.5, 0.75, 0.9].

%With curiosity, we conducted further experiments, and found that the reviewer was right about this: the above Deep $k$-means routine will frequently lead to many small but non-zero cluster centers. 

% We thus implement a similar sparsity-promoting feature for Deep $k$-Means. 

%We empirically observe such a sparsity-promotion operation to marginally decrease compression performance as it restricts the flexibility of setting centroids. However, it brings in both run-time accelerations as well as energy benefits. We thus include sparsity promotion as default for Deep $k$-means hereinafter. 

%This is also the only experiment in this paper that we varied and fine-tuned $p$ per layer, as we empirically found Wide ResNet more sensitive to weight sparsity than other tested models. The best set of layer-wise $p$ values we found here is: [0, 0.3, 0.4, 0.5, 0.4, 0.4, 0.5, 0.5, 0.5 , 0.5 , 0.5, 0.6, 0.9, 0.5, 0.75 0.9]. 

Table \ref{WRN} demonstrates the superiority of \textit{Deep $k$-Means} (with the above-described sparsity promotion) over \cite{ullrich2017soft}, by comparing their top-1 accuracy drops: 1.63\% versus 2.02 \%, at CR = 45. We further display the results at CR = 47 and 50, with a smooth accuracy decrease. 

%\footnote{We notice that the authors of \cite{ullrich2017soft} did not exactly reproduce the result of \cite{zagoruyko2016wide}, while we managed to do so in our experiments. That means our implemented uncompressed Wide ResNet to have higher accuracy than the one used in \cite{ullrich2017soft}. In view of that, we think the comparison of the accuracy loss ($\Delta$) to be more fair than comparing absolute accuracy values between the two works.}

%Second, Wide ResNet appears to be tremendously sensitive to directly quantizing filters using \textit{Deep $k$-Means WP}: the accuracies are completely deteriorated under all CRs evaluated. \textit{Deep $k$-Means} shows the ability to overcome this problem and to remedy accuracies to a surprising extent, validating the necessity of our proposed re-training particularly in compressing large modern CNNs. As an additional interesting observation, CR = 45 might be right around the border for compressing Wide ResNet: a larger CR than 45 (e.g., 45.5 and 48) seems to cause rapid accuracy degradations even for our \textit{Deep $k$-Means}. 

\subsubsection{Evaluation with GoogleNet on ImageNet}

We finally evaluate \textit{Deep $k$-Means} on the GoogleNet \cite{szegedy2015going} trained with the ImageNet ILSVRC12 dataset \cite{russakovsky2015imagenet}. We use single center crop during testing, and evaluate the performance based on the \textit{top-1} and \textit{top-5} accuracy drops on the validation set, compared to the uncompressed baseline whose \textit{top-1} accuracy is 69.76\% and \textit{top-5} 89.63\%. We include two comparison methods: one-shot network compression \cite{kim2015compression}, and low-rank regularization \cite{tai2015convolutional}. According to Table \ref{Google}, \textit{Deep $k$-Means} proves to scale well on large models/datasets, and achieves significantly better results over the two baselines: its compression at CR $\le$ 3 is almost lossless, with top-5 errors again observed to slightly increase after compression. The GoogleNet compression performance is found to deteriorate quickly when CR $>$ 4. 

%with more competitive performance achieved than the two baselines. 

%\vspace{-1em}
\begin{table}[htp!]
\centering
\begin{tabular}{llll}
    \toprule
    Model    & $\Delta$\textsuperscript{$\dagger$} \% & $\Delta$\textsuperscript{$\ddagger$} \% & CR  \\
    \midrule
    One-shot \cite{kim2015compression} & N/A & -$0.24$ & $1.28$ \\
    Low-rank \cite{tai2015convolutional} & N/A & -$0.42$ & $2.84$ \\
    \hline
    %Deep $k$-Means WR & -$53.35$ & $40$ \\
    Deep $k$-Means WR & -$1.22$ & -$0.65$ & $1.5$ \\
    Deep $k$-Means WR & -$3.7$ & -$2.46$ & $2$ \\
    Deep $k$-Means WR & -$13.72$ & -$10.05$ & $3$ \\
        Deep $k$-Means WR & -$48.95$ & -$48.82$ & $4$ \\
    \hline
    %Deep $k$-Means & -$1.42$ & $40$ \\
    Deep $k$-Means & -$0.26$ & $0.00$ & $1.5$ \\
    Deep $k$-Means & -$0.17$ & +$0.06$ & $2$ \\
    Deep $k$-Means & -$0.36$ & +$0.03$ & $3$ \\
    Deep $k$-Means & -$1.95$ & -$1.14$ & $4$ \\
    \bottomrule
  \end{tabular}
\caption{Compressing GoogLeNet on ILSVRC12 (\textsuperscript{$\dagger$} and \textsuperscript{$\ddagger$} are top-1 and top-5 accuracies respectively).}
\vspace{-2em}
\label{Google}
\end{table}

%GoogleNet
% [1] Compression of Deep CNNs for Fast Low Power Mobile Apps,
% ICLR’16
% - CR 1.28, top-5 error + 0.24 (%)
% [2] CNNs with low-rank regularization, ICLR’16
% - CR 2.84, + 0.42
% Ours:
% - CR 2, + 0.04
% - CR 3, + 0.47

% \begin{figure*}[htbp]
% \begin{center}
% \begin{tabular}{cccc}
% \hspace {-0.2 cm }
%   \includegraphics[width=0.24\textwidth]{images/conv1_retrain_computational_cost.png} &\hspace {-1.0cm }
%   \includegraphics[width=0.24\textwidth]{images/conv2_retrain_computational_cost.png} &\hspace {-1.0cm }
%   \includegraphics[width=0.24\textwidth]{images/conv1_retrain_weight_representational_cost.png} &\hspace {-1.0cm }
%   \includegraphics[width=0.24\textwidth]{images/conv2_retrain_weight_representational_cost.png} \\
%   (a) Conv1 Layer  &  (b) Conv2 Layer & (c) Conv1 Layer  &  (d) Conv2 Layer  \\
% \end{tabular}
% \end{center}
% \caption{Accuracy}
% \label{computational_cost}
% \end{figure*}

% \begin{figure}[htbp]
% \centering
% \begin{minipage}{0.45\textwidth}
% \centering{
% \hspace {-0.2 cm }
%   \includegraphics[width=0.24\textwidth]{images/conv1_retrain_computational_cost.png}
%   \includegraphics[width=0.24\textwidth]{images/conv2_retrain_computational_cost.png}
%   \includegraphics[width=0.24\textwidth]{images/conv1_retrain_weight_representational_cost.png}
%   \includegraphics[width=0.24\textwidth]{images/conv2_retrain_weight_representational_cost.png}
  
% }\end{minipage}
% \caption{Energy Consumption Comparison using \textit{Deep $k$-Means}}
% %\vspace{-1em}
% \label{ablation}
% \end{figure}

%\vspace{-1em}
\subsection{Comparison on Energy-Aware Metrics}
%\textcolor{red}{To be re-done by Yingyan}
%\vspace{-0.1em}
\subsubsection{Energy-Aware Metrics Verification} 
\vspace{-0.3em}
%Cluster Rate for AlexNet: [50, 70, 75, 80, 85, 90, 95, 99]
%Cluster Rate for GoogLeNet: [67, 82, 95, 98.8]

We first evaluate our energy-aware metrics by comparing its estimated energy consumption with that of the tool in \cite{yang2017designing}. Note that the unit of energy: 1) in \cite{yang2017designing} is normalized in terms of number of MAC operations while the computational cost in Eq. \eqref{eq: computational cost} is normalized in terms of number of FAs; and 2) for the representational cost in  Eqs. \eqref{eq: weight representational cost} and \eqref{eq: activation representational cost} is different from that of the computational cost in Eq. \eqref{eq: computational cost}. Therefore, we first normalize the representational cost in terms of the computational cost assuming that a global on-chip buffer is employed, implying that the representational cost of a MAC is about $6$ times that of performing a MAC computation \cite{Eyeriss-isca}. This normalized representational cost is then added to the computational cost to obtain our total energy. Lastly, we normalize this total energy in terms of the number of MACs to be the same as that of \cite{Eyeriss-isca}. 

We calculate the coefficient of determination ($R^2$), between the estimated energy consumptions using our proposed metrics, and using the tool in \cite{yang2017designing}, of the same compressed models. We use \textit{Deep $k$-means} to compress both AlexNet and GoogLeNet_v1 \footnote{For both networks, we employ our proposed methods in convolutional layers only. For AlexNet, we only quantize weight and activation to 8 bit and to 16 bit in fully-connected layers, respectively. For GooLeNet_v1, we use the one with global average pooling, which has no fully-connected layer.}, which are the \textit{only two CNN models} currently supported by \cite{yang2017designing}. 
%For AlexNet, we did not compress the fully-connected layer. we use the GooLeNet_v1 with global average pooling, where there is no fully-connected layer. 
The energy consumptions are estimated as
we choose the \textit{cluster rate} to vary between: (AlexNet)
% [50, 70, 75, 80, 85, 90, 95, 99] and [67, 82, 95, 98.8], respectively,
[0.5, 0.3, 0.25, 0.2, 0.15, 0.1, 0.05, 0.01], and (GoogLeNet_v1) [0.33, 0.18, 0.05, 0.012], respectively, to ensure negligible accuracy loss. We have $R^2$ 
%between the resulting energy using our metrics and the tool in \cite{yang2017designing} 
to be \textbf{0.9931} for AlexNet, and \textbf{0.9675} for GoogLeNet_v1, suggesting the estimated energy consumptions using our proposed metrics to be \textit{strongly linearly correlated} with the results extrapolated from actual hardware measurements \cite{yang2017designing}. Yet different from their tool, our metrics are generally applicable to any CNN.

%summarized in Table \ref{R2_score}. We can see that $R^2$ is equal to $0.9931$ and $0.9675$, respectively, for the two considered CNNs, 

%indicating a good match between the estimated energy using our proposed metrics and the tool derived from actual hardware measurements \cite{yang2017designing}. 

%For example, a $B_{\mathbf{w}}$ and $B_{\mathbf{x}}$ MAC operation corresponds to about $B_{\mathbf{w}}\times B_{\mathbf{x}} + max{B_{\mathbf{w}}, B_{\mathbf{x}}}$ FAs. 
% \vspace{-2em} 
% \begin{table}[htp!]
% \centering
% \begin{tabular}{lll}
%     \toprule
%     Model    & $R^2$ score  \\
%     \midrule
%     AlexNet & $0.9931$ \\
%     GoogLenet & $0.9675$ \\
%     \bottomrule
%   \end{tabular}
% \vspace{-0.7em}
% \caption{Comparing the coefficient of determination, denoted as $R^2$, between the estimated energy consumption from our metrics and the tool in \cite{yang2017designing}  in the cases of AlexNet and GoogLeNet_v1.}
% \vspace{-2em}
% \label{R2_score}
% \end{table}

\vspace{-0.3em}
\subsubsection{Comparison with Greedy Filter Pruning}
An ideal CNN model to be deployed on resource-constrained platforms should simultaneously possess compact model size and low energy cost. 
%while greedy filter pruning and soft weight-sharing both give inferior but still competitive CR results. 
%To evaluate Deep $k$-Means on the potential energy cost reduction in the resulted CNNs, we compare Deep $k$-Means against the four baselines in terms of the computational and representational costs defined in Eqs. \eqref{eq: computational cost}, \eqref{eq: weight representational cost} and \eqref{eq: activation representational cost}. 
In general, the resulting computational/representational cost reduction via compression depends on how the network is trimmed, i.e., which parts of the network is compressed. Conceptually, we point out that different compression schemes (e.g., parameter-sharing versus pruning) will affect the analysis of the computational and representational costs defined in Eqs. \eqref{eq: computational cost}, \eqref{eq: weight representational cost} and \eqref{eq: activation representational cost}. 
First, the weight representational cost is directly proportional to CR in both parameter-sharing (e.g. \textit{Deep $k$-Means} and soft weight-sharing) and pruning (e.g. CAR) cases, because they both in effect can reduce the term $\left|\mathcal{W}\right|$ in \eqref{eq: weight representational cost}. Second, the activation representational cost is proportional to CR for the case of weight pruning, but is independent of CR if the compression is done by weight sharing. This is because weight pruning results in skipping of the corresponding computations and thus can reduce the number of times that the corresponding activations are used (i.e., $\mathcal{W}$ in \eqref{eq: activation representational cost}), whereas there is no computation or connection skipping in the case of weight-sharing. Third, the computational cost is again proportional to CR for weight pruning; yet it would become input-dependent when it comes to weight sharing. 
%depend on the positions of shared weights for the case of weight sharing. 
Specifically, only when all the weights corresponding to the same input/activation are shared, the computational cost reduction ratio becomes equal to CR. 

% Stacked Configure
\begin{figure}[htbp]
\vspace{-0.3em}
\centering
\begin{minipage}{0.40\textwidth}
\centering\subfigure[Comparison in the first convolutional layer]{
\includegraphics[width=\textwidth, height=0.2\textheight]{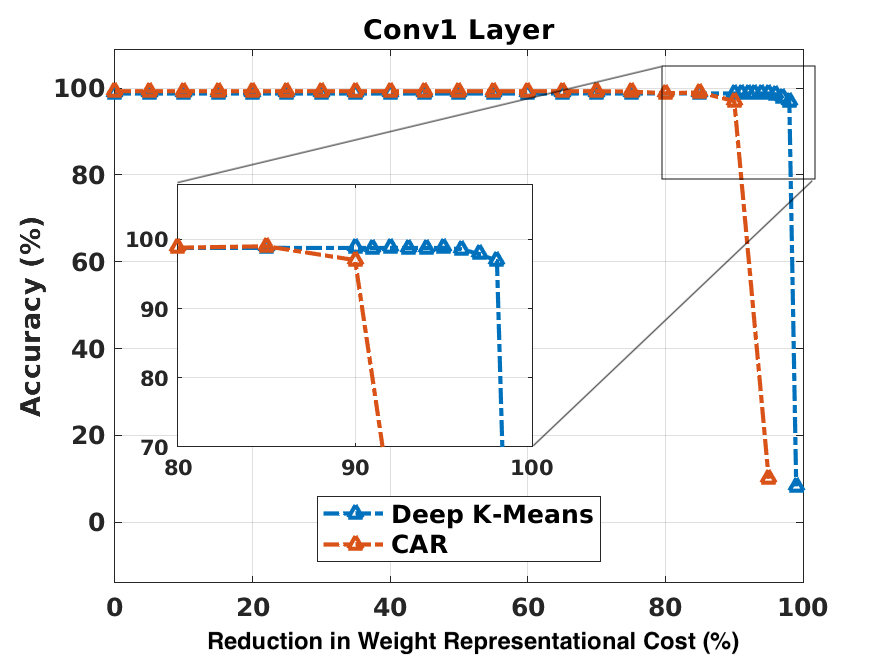}
}\end{minipage}
\begin{minipage}{0.40\textwidth}
% \vspace{-0.5em}
\centering\subfigure[Comparison in the second convolutional layer]{
\includegraphics[width=\textwidth, height=0.2\textheight]{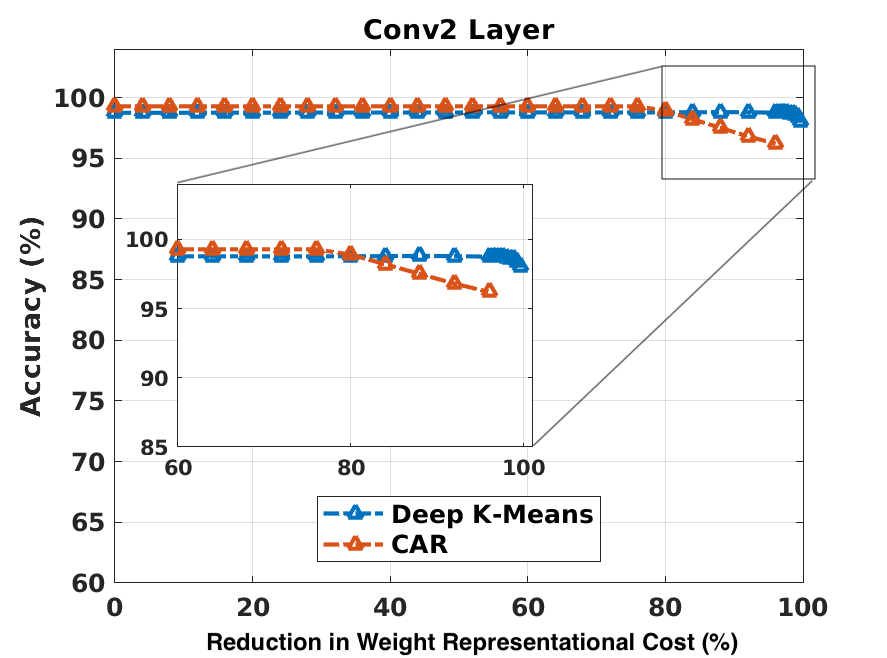}
}\end{minipage}
\vspace{-1em}
\caption{Comparison between \textit{Deep $k$-Means} and CAR, in terms of the ratio between weight representational cost reduction.}
\vspace{-0.3em}
\label{greedy2}
\end{figure}

% Side by Side Configure
% \begin{figure*}[htbp]
% \centering
% \begin{minipage}{0.4\textwidth}
% \centering\subfigure[]{
% \includegraphics[width=\textwidth]{images/conv1_retrain_weight_representational_cost.png}
% }\end{minipage}
% \begin{minipage}{0.4\textwidth}
% \centering\subfigure[]{
% \includegraphics[width=\textwidth]{images/conv2_retrain_weight_representational_cost.png}
% }\end{minipage}
% \vspace{-1em}
% \caption{Comparison between \textit{Deep $k$-Means} and CAR, in terms of the ratio between weight representational cost reduction over the original CNN model.}
% \vspace{-0.5em}
% \label{greedy2}
% \end{figure*}
\textit{Deep $k$-Means} has constantly obtained the best CR performance among the aforementioned baselines. To provide a concrete example, we choose the CAR (with re-training) baseline in \cite{abbasi2017structural}, which produces slightly inferior but still competitive CR results, and discuss its potential energy efficiency improvement compared with \textit{Deep $k$-Means}. The same layer-wise compression setting in Section \ref{greedy_section} is adopted, for the first two convolutional layers of LeNet-5.
%in terms of the computational and representational costs defined in Eqs. \eqref{eq: computational cost}, \eqref{eq: weight representational cost} and \eqref{eq: activation representational cost}. 
A similar analysis could be done for other methods too.

%Before we compare the impact of Deep $k$-Means and the baselines on the computational and representational costs, we discuss the relationship between CR and these energy-aware metrics to provide conceptual insights. 

%First, the weight representational cost is directly proportional to CR in both weight-sharing (e.g. Deep $k$-Means and soft weight-sharing) and weight pruning (e.g. CAR) cases, because they both in effect can reduce the term $\left|\mathcal{X}\right|$ in \eqref{eq: weight representational cost}. Second, the activation representational cost is directly proportional to CR for the case of compression via weight pruning, and is independent of CR for the case of compression via weight sharing. This is because weight pruning results in skipping of the corresponding computations and thus can reduce the number of times that the corresponding activations are used (i.e., $N_{\mathbf{x}}$ in \eqref{eq: activation representational cost}), whereas there is no computation or connection skipping in the case of weight-sharing. Third, the computational cost is directly proportional to CR when using the compression is performed through weight pruning, and it would depends on the positions of shared weights for the case of compression through weight sharing. Specifically, only when all the weights corresponding to the same input/activation can be shared, the ratio on the computational cost reduction over the original model is equal to CR. 

%We then calculate the computational and representational costs of \textit{Deep $k$-Means} and CAR to provide an concrete example. 
\textit{Deep $k$-Means} compresses the network via weight sharing, whereas CAR relies on weight pruning.  The accuracy versus CR comparison (i.e., Figure \ref{greedy}) in Section \ref{greedy_section} shows that \textit{Deep $k$-Means} is clearly superior to CAR at high layer-wise CRs, for either of the two convolutional layers. The potential energy consumption comparison between \textit{Deep $k$-Means} and CAR in terms of the three metrics are as follows\footnote{We are unable to directly verify the total energy consumption of CAR using our metrics due to the lack of their model parameters or pre-trained model.}. 

First, \textit{Deep $k$-Means} will achieve higher weight representational cost reduction since it mostly offers higher CR with the same or even better accuracy, in particular at high CRs. Figure \ref{greedy2} (a) and (b) compares the accuracy versus weight representational cost reduction ratio of \textit{Deep $k$-Means} and CAR for compressing the first and second convolutional layers in LeNet 5, respectively\footnote{We did not consider the cost of weight assignment indexes as it is negligible due to achievable high CR.}. We observe that \textit{Deep $k$-Means} achieves about 10\% and 17.2\% higher weight representational cost reduction, respectively, compared to CAR when compressing the first and second layers, without incurring accuracy loss. Second, CAR always outperforms \textit{Deep $k$-Means} in terms of activation representational cost, because it removes filters and thus reduces the numbers of feature maps. In theory,  CAR can achieve up to (layer-wise) ``CR times'' better activation representational cost reduction ratio than \textit{Deep $k$-Means}. Third, the achievable computational cost reduction by \textit{Deep $k$-Means} is either smaller or equal to that of CAR, depending on the inputs. 

%On the other hand, 
%in the first and second convolutional layers of LeNet 5. % (please fill in the concrete ). In conclusion, although Deep $k$-slightly outperform CAR in terms of CR, whether this lead to better potential energy cost reduction depends on the percentage of computational and representational costs, which in turns is determined by the CNN model and the dataflow.

%the equation is a reasonable estimate, but may not perfect reflect energy. We note \cite{yang2017designing} ...

% \subsection{More Discussions}
% [Optional]:
% ablation study on $k$ with NIN -> non-uniform $k$ may further improve results \\
% even Le-Net or AlexNet: no touch fc, just conv, compare energy only\\
%why it fails on fc, how it can be improved by, e.g., using other methods?\\

\vspace{-0.5em}
\section{Conclusion and Discussions}

%\vspace{-0.5em}
This paper proposes \textit{Deep $k$-Means}, a retraining-then-parameter-sharing pipeline for compressing convolutional layers in deep CNNs. A novel spectrally relaxed $k$-means regularization is derived to make hard assignments of convolutional layer weights to learned cluster centers during re-training. \textit{Deep $k$-Means} demonstrates clear superiority over several recently-proposed competitive methods, in terms of both compression ratio and energy efficiency. 
%It sets the new state-of-the-art result benchmark on the Wide ResNet model. In particular, we are able to achieve 45 times compression of the state-of-the-art \textbf{Wide ResNet} model with only 0.51\% loss of top-1 accuracy. 
Our future work will exploit more adaptive cluster rates for different layers instead of the current uniform scheme. Based on our proposed metrics, we also aim to incorporate more energy-aware regularizations into \textit{Deep $k$-Means} for direct minimization of energy consumptions.

\section*{Acknowledgements}
We would like to thank all anonymous reviewers for their tremendously useful comments to help improve our work. We acknowledge the inspiring discussions with Dr. Yang Zhang at IBM Watson and Dr. Edwin Park at Qualcomm Technologies, Inc. We acknowledge the Texas A\&M and Rice High Performance Research Computing for providing a part of the computing resources used in this research.

\bibliography{example_paper}

\begin{thebibliography}{42}
\providecommand{\natexlab}[1]{#1}
\providecommand{\url}[1]{\texttt{#1}}
\expandafter\ifx\csname urlstyle\endcsname\relax
  \providecommand{\doi}[1]{doi: #1}\else
  \providecommand{\doi}{doi: \begingroup \urlstyle{rm}\Url}\fi

\bibitem[Abbasi-Asl \& Yu(2017)Abbasi-Asl and Yu]{abbasi2017structural}
Abbasi-Asl, Reza and Yu, Bin.
\newblock Structural compression of convolutional neural networks based on
  greedy filter pruning.
\newblock \emph{arXiv preprint arXiv:1705.07356}, 2017.

\bibitem[Basu et~al.(2002)Basu, Banerjee, and Mooney]{basu2002semi}
Basu, Sugato, Banerjee, Arindam, and Mooney, Raymond.
\newblock Semi-supervised clustering by seeding.
\newblock In \emph{In Proceedings of 19th International Conference on Machine
  Learning (ICML-2002}. Citeseer, 2002.

\bibitem[Bhattacharya \& Lane(2016)Bhattacharya and
  Lane]{bib:bhattacharya2016:CD-ROM}
Bhattacharya, Sourav and Lane, Nicholas~D.
\newblock Sparsification and separation of deep learning layers for constrained
  resource inference on wearables.
\newblock In \emph{Proceedings of SenSys}, 2016.

\bibitem[Changpinyo et~al.(2017)Changpinyo, Sandler, and
  Zhmoginov]{bib:ICLR2017:soravit}
Changpinyo, Soravit, Sandler, Mark, and Zhmoginov, Andrey.
\newblock The power of sparsity in convolutional neural networks.
\newblock \emph{arXiv preprint arXiv:1702.06257}, 2017.

\bibitem[Chen et~al.(2015)Chen, Wilson, Tyree, Weinberger, and
  Chen]{DBLP:journals/corr/ChenWTWC15}
Chen, Wenlin, Wilson, James~T., Tyree, Stephen, Weinberger, Kilian~Q., and
  Chen, Yixin.
\newblock Compressing neural networks with the hashing trick.
\newblock \emph{CoRR}, abs/1504.04788, 2015.

\bibitem[Chen et~al.(2016{\natexlab{a}})Chen, Wilson, Tyree, Weinberger, and
  Chen]{chen2016compressing}
Chen, Wenlin, Wilson, James, Tyree, Stephen, Weinberger, Kilian~Q, and Chen,
  Yixin.
\newblock Compressing convolutional neural networks in the frequency domain.
\newblock In \emph{Proceedings of the 22nd ACM SIGKDD International Conference
  on Knowledge Discovery and Data Mining}, pp.\  1475--1484. ACM,
  2016{\natexlab{a}}.

\bibitem[Chen et~al.(2016{\natexlab{b}})Chen, Emer, and Sze]{Eyeriss-isca}
Chen, Y.~H., Emer, J., and Sze, V.
\newblock Eyeriss: A spatial architecture for energy-efficient dataflow for
  convolutional neural networks.
\newblock In \emph{2016 ACM/IEEE 43rd Annual International Symposium on
  Computer Architecture (ISCA)}, pp.\  367--379, June 2016{\natexlab{b}}.

\bibitem[Cheng et~al.(2017)Cheng, Wang, Zhou, and Zhang]{cheng2017survey}
Cheng, Yu, Wang, Duo, Zhou, Pan, and Zhang, Tao.
\newblock A survey of model compression and acceleration for deep neural
  networks.
\newblock \emph{arXiv preprint arXiv:1710.09282}, 2017.

\bibitem[Cun et~al.(1990)Cun, Denker, and Solla]{Cun:1990:OBD:109230.109298}
Cun, Yann~Le, Denker, John~S., and Solla, Sara~A.
\newblock Advances in neural information processing systems 2.
\newblock chapter Optimal Brain Damage, pp.\  598--605. 1990.
\newblock ISBN 1-55860-100-7.

\bibitem[Denil et~al.(2013)Denil, Shakibi, Dinh, De~Freitas,
  et~al.]{denil2013predicting}
Denil, Misha, Shakibi, Babak, Dinh, Laurent, De~Freitas, Nando, et~al.
\newblock Predicting parameters in deep learning.
\newblock In \emph{Advances in neural information processing systems}, pp.\
  2148--2156, 2013.

\bibitem[Garipov et~al.(2016)Garipov, Podoprikhin, Novikov, and
  Vetrov]{garipov2016ultimate}
Garipov, Timur, Podoprikhin, Dmitry, Novikov, Alexander, and Vetrov, Dmitry.
\newblock Ultimate tensorization: compressing convolutional and fc layers
  alike.
\newblock \emph{arXiv preprint arXiv:1611.03214}, 2016.

\bibitem[Girshick et~al.(2013)Girshick, Donahue, Darrell, and Malik]{Object}
Girshick, Ross~B., Donahue, Jeff, Darrell, Trevor, and Malik, Jitendra.
\newblock Rich feature hierarchies for accurate object detection and semantic
  segmentation.
\newblock volume abs/1311.2524, 2013.

\bibitem[Gong et~al.(2014{\natexlab{a}})Gong, Liu, Yang, and
  Bourdev]{gong2014compressing}
Gong, Yunchao, Liu, Liu, Yang, Ming, and Bourdev, Lubomir.
\newblock Compressing deep convolutional networks using vector quantization.
\newblock \emph{arXiv preprint arXiv:1412.6115}, 2014{\natexlab{a}}.

\bibitem[Gong et~al.(2014{\natexlab{b}})Gong, Liu, Yang, and
  Bourdev]{DBLP:journals/corr/GongLYB14}
Gong, Yunchao, Liu, Liu, Yang, Ming, and Bourdev, Lubomir~D.
\newblock Compressing deep convolutional networks using vector quantization.
\newblock \emph{CoRR}, abs/1412.6115, 2014{\natexlab{b}}.

\bibitem[Han et~al.(2015)Han, Pool, Tran, and
  Dally]{DBLP:journals/corr/HanPTD15}
Han, Song, Pool, Jeff, Tran, John, and Dally, William~J.
\newblock Learning both weights and connections for efficient neural networks.
\newblock \emph{CoRR}, abs/1506.02626, 2015.

\bibitem[Han et~al.(2016)Han, Mao, and Dally]{bib:arXiv2015:han}
Han, Song, Mao, Huizi, and Dally, William~J.
\newblock Deep compression: Compressing deep neural networks with pruning,
  trained quantization and huffman coding.
\newblock In \emph{Proceedings of ICLR}, 2016.

\bibitem[Hanson \& Pratt(1989)Hanson and Pratt]{NIPS1988_156}
Hanson, Stephen~Jose and Pratt, Lorien~Y.
\newblock Comparing biases for minimal network construction with
  back-propagation.
\newblock In Touretzky, D.~S. (ed.), \emph{Advances in Neural Information
  Processing Systems 1}, pp.\  177--185. Morgan-Kaufmann, 1989.

\bibitem[Hassibi \& Stork(1993)Hassibi and Stork]{NIPS1992_647}
Hassibi, Babak and Stork, David~G.
\newblock Second order derivatives for network pruning: Optimal brain surgeon.
\newblock In Hanson, S.~J., Cowan, J.~D., and Giles, C.~L. (eds.),
  \emph{Advances in Neural Information Processing Systems 5}, pp.\  164--171.
  Morgan-Kaufmann, 1993.

\bibitem[He et~al.(2014)He, Fan, Qian, Tan, and Yu]{He}
He, T., Fan, Y., Qian, Y., Tan, T., and Yu, K.
\newblock Reshaping deep neural network for fast decoding by node-pruning.
\newblock In \emph{2014 IEEE International Conference on Acoustics, Speech and
  Signal Processing (ICASSP)}, pp.\  245--249, May 2014.

\bibitem[Howard et~al.(2017)Howard, Zhu, Chen, Kalenichenko, Wang, Weyand,
  Andreetto, and Adam]{bib:arXiv2017:Howard}
Howard, Andrew~G, Zhu, Menglong, Chen, Bo, Kalenichenko, Dmitry, Wang, Weijun,
  Weyand, Tobias, Andreetto, Marco, and Adam, Hartwig.
\newblock Mobilenets: Efficient convolutional neural networks for mobile vision
  applications.
\newblock \emph{arXiv preprint arXiv:1704.04861}, 2017.

\bibitem[Iandola et~al.(2016)Iandola, Han, Moskewicz, Ashraf, Dally, and
  Keutzer]{bib:iandola2016:arxiv}
Iandola, Forrest~N, Han, Song, Moskewicz, Matthew~W, Ashraf, Khalid, Dally,
  William~J, and Keutzer, Kurt.
\newblock Squeezenet: Alexnet-level accuracy with 50x fewer parameters and< 0.5
  mb model size.
\newblock \emph{arXiv preprint arXiv:1602.07360}, 2016.

\bibitem[Kim et~al.(2015)Kim, Park, Yoo, Choi, Yang, and
  Shin]{kim2015compression}
Kim, Yong-Deok, Park, Eunhyeok, Yoo, Sungjoo, Choi, Taelim, Yang, Lu, and Shin,
  Dongjun.
\newblock Compression of deep convolutional neural networks for fast and low
  power mobile applications.
\newblock \emph{arXiv preprint arXiv:1511.06530}, 2015.

\bibitem[Krizhevsky \& Hinton(2009)Krizhevsky and
  Hinton]{krizhevsky2009learning}
Krizhevsky, Alex and Hinton, Geoffrey.
\newblock Learning multiple layers of features from tiny images.
\newblock 2009.

\bibitem[Krizhevsky et~al.(2012)Krizhevsky, Sutskever, and Hinton]{AlexNet}
Krizhevsky, Alex, Sutskever, Ilya, and Hinton, Geoffrey~E.
\newblock Imagenet classification with deep convolutional neural networks.
\newblock In Pereira, F., Burges, C. J.~C., Bottou, L., and Weinberger, K.~Q.
  (eds.), \emph{Advances in Neural Information Processing Systems 25}, pp.\
  1097--1105. Curran Associates, Inc., 2012.

\bibitem[Lane et~al.(2016)Lane, Bhattacharya, Georgiev, Forlivesi, Jiao,
  Qendro, and Kawsar]{bib:lane2016:IPSN}
Lane, Nicholas~D, Bhattacharya, Sourav, Georgiev, Petko, Forlivesi, Claudio,
  Jiao, Lei, Qendro, Lorena, and Kawsar, Fahim.
\newblock Deepx: A software accelerator for low-power deep learning inference
  on mobile devices.
\newblock In \emph{Proceedings of IPSN}, 2016.

\bibitem[Li et~al.(2016)Li, Kadav, Durdanovic, Samet, and
  Graf]{DBLP:journals/corr/LiKDSG16}
Li, Hao, Kadav, Asim, Durdanovic, Igor, Samet, Hanan, and Graf, Hans~Peter.
\newblock Pruning filters for efficient convnets.
\newblock \emph{CoRR}, abs/1608.08710, 2016.

\bibitem[Lin et~al.(2014)Lin, Chen, and Yan]{bib:lin2013:NIN}
Lin, Min, Chen, Qiang, and Yan, Shuicheng.
\newblock Network in network.
\newblock In \emph{Proceedings of ICLR}, 2014.

\bibitem[Lin et~al.(2016)Lin, Zhang, and Shanbhag]{RD-SEC}
Lin, Y., Zhang, S., and Shanbhag, N.~R.
\newblock Variation-tolerant architectures for convolutional neural networks in
  the near threshold voltage regime.
\newblock In \emph{2016 IEEE International Workshop on Signal Processing
  Systems (SiPS)}, pp.\  17--22, Oct 2016.

\bibitem[Lin et~al.(2017)Lin, Sakr, Kim, and Shanbhag]{ISCAS_PredictiveNet}
Lin, Yingyan, Sakr, Charbel, Kim, Yongjune, and Shanbhag, Naresh.
\newblock Predictivenet: An energy-efficient convolutional neural network via
  zero prediction.
\newblock In \emph{Proceedings of ISCAS}, 2017.

\bibitem[Parashar et~al.(2017)Parashar, Rhu, Mukkara, Puglielli, Venkatesan,
  Khailany, Emer, Keckler, and Dally]{parashar2017scnn}
Parashar, Angshuman, Rhu, Minsoo, Mukkara, Anurag, Puglielli, Antonio,
  Venkatesan, Rangharajan, Khailany, Brucek, Emer, Joel, Keckler, Stephen~W,
  and Dally, William~J.
\newblock Scnn: An accelerator for compressed-sparse convolutional neural
  networks.
\newblock In \emph{Proceedings of the 44th Annual International Symposium on
  Computer Architecture}, pp.\  27--40. ACM, 2017.

\bibitem[Russakovsky et~al.(2015)Russakovsky, Deng, Su, Krause, Satheesh, Ma,
  Huang, Karpathy, Khosla, Bernstein, et~al.]{russakovsky2015imagenet}
Russakovsky, Olga, Deng, Jia, Su, Hao, Krause, Jonathan, Satheesh, Sanjeev, Ma,
  Sean, Huang, Zhiheng, Karpathy, Andrej, Khosla, Aditya, Bernstein, Michael,
  et~al.
\newblock Imagenet large scale visual recognition challenge.
\newblock \emph{International Journal of Computer Vision}, 115\penalty0
  (3):\penalty0 211--252, 2015.

\bibitem[Sakr et~al.(2017)Sakr, Kim, and Shanbhag]{pmlr-v70-sakr17a}
Sakr, Charbel, Kim, Yongjune, and Shanbhag, Naresh.
\newblock Analytical guarantees on numerical precision of deep neural networks.
\newblock In Precup, Doina and Teh, Yee~Whye (eds.), \emph{Proceedings of the
  34th International Conference on Machine Learning}, volume~70 of
  \emph{Proceedings of Machine Learning Research}, pp.\  3007--3016,
  International Convention Centre, Sydney, Australia, 06--11 Aug 2017. PMLR.

\bibitem[Shi et~al.(2016)Shi, Cao, Zhang, Li, and Xu]{Edge}
Shi, W., Cao, J., Zhang, Q., Li, Y., and Xu, L.
\newblock Edge computing: Vision and challenges.
\newblock volume~3, pp.\  637--646, Oct 2016.

\bibitem[Srinivas \& Babu(2015)Srinivas and
  Babu]{DBLP:journals/corr/SrinivasB15}
Srinivas, Suraj and Babu, R.~Venkatesh.
\newblock Data-free parameter pruning for deep neural networks.
\newblock \emph{CoRR}, abs/1507.06149, 2015.

\bibitem[Szegedy et~al.(2015)Szegedy, Liu, Jia, Sermanet, Reed, Anguelov,
  Erhan, Vanhoucke, Rabinovich, et~al.]{szegedy2015going}
Szegedy, Christian, Liu, Wei, Jia, Yangqing, Sermanet, Pierre, Reed, Scott,
  Anguelov, Dragomir, Erhan, Dumitru, Vanhoucke, Vincent, Rabinovich, Andrew,
  et~al.
\newblock Going deeper with convolutions.
\newblock CVPR, 2015.

\bibitem[Tai et~al.(2015)Tai, Xiao, Zhang, Wang, et~al.]{tai2015convolutional}
Tai, Cheng, Xiao, Tong, Zhang, Yi, Wang, Xiaogang, et~al.
\newblock Convolutional neural networks with low-rank regularization.
\newblock \emph{arXiv preprint arXiv:1511.06067}, 2015.

\bibitem[Taigman et~al.(2014)Taigman, Yang, Ranzato, and Wolf]{DeepFace}
Taigman, Yaniv, Yang, Ming, Ranzato, Marc'Aurelio, and Wolf, Lior.
\newblock Deepface: Closing the gap to human-level performance in face
  verification.
\newblock In \emph{Proceedings of the IEEE conference on computer vision and
  pattern recognition}, pp.\  1701--1708, 2014.

\bibitem[Ullrich et~al.(2017)Ullrich, Meeds, and Welling]{ullrich2017soft}
Ullrich, Karen, Meeds, Edward, and Welling, Max.
\newblock Soft weight-sharing for neural network compression.
\newblock \emph{arXiv preprint arXiv:1702.04008}, 2017.

\bibitem[Wang et~al.(2015)Wang, Yang, Jin, Shechtman, Agarwala, Brandt, and
  Huang]{wang2015deepfont}
Wang, Zhangyang, Yang, Jianchao, Jin, Hailin, Shechtman, Eli, Agarwala, Aseem,
  Brandt, Jonathan, and Huang, Thomas~S.
\newblock Deepfont: Identify your font from an image.
\newblock In \emph{Proceedings of the 23rd ACM international conference on
  Multimedia}, pp.\  451--459. ACM, 2015.

\bibitem[Yang et~al.(2017)Yang, Chen, and Sze]{yang2017designing}
Yang, Tien-Ju, Chen, Yu-Hsin, and Sze, Vivienne.
\newblock Designing energy-efficient convolutional neural networks using
  energy-aware pruning.
\newblock \emph{arXiv preprint}, 2017.

\bibitem[Zagoruyko \& Komodakis(2016)Zagoruyko and
  Komodakis]{zagoruyko2016wide}
Zagoruyko, Sergey and Komodakis, Nikos.
\newblock Wide residual networks.
\newblock \emph{arXiv preprint arXiv:1605.07146}, 2016.

\bibitem[Zha et~al.(2002)Zha, He, Ding, Gu, and Simon]{zha2002spectral}
Zha, Hongyuan, He, Xiaofeng, Ding, Chris, Gu, Ming, and Simon, Horst~D.
\newblock Spectral relaxation for k-means clustering.
\newblock In \emph{Advances in neural information processing systems}, pp.\
  1057--1064, 2002.

\end{thebibliography}
\bibliographystyle{icml2018}

\end{document}